\documentclass[letterpaper, 10 pt, conference]{ieeeconf}  

\usepackage{graphicx}
\usepackage{amsmath}
\usepackage{subcaption}
\usepackage{url}
\usepackage{cooltooltips}
\usepackage{fixme}
\usepackage{hyperref}

\title{\LARGE \bf
Self-Stabilizing Self-Assembly
}

\author{\authorblockN{Martin J\'{i}lek$^{1}$, Kate\v{r}ina Str\'{a}nsk\'{a}$^{1}$, Michael Somr$^{2}$, Miroslav Kulich$^{1}$, Jan Zeman$^{2}$, and Libor P\v{r}eu\v{c}il$^{1}$}\authorblockA{$^{1}$Czech Institute of Informatics, Robotics and Cybernetics\\Czech Technical University in Prague\\E-mail:
		\{martin.jilek.2, miroslav.kulich, katerina.stranska, libor.preucil\}@cvut.cz
	}
	\authorblockA{$^{2}$Faculty of Civil Engineering\\Czech Technical University in Prague\\E-mail: michael.somr@fsv.cvut.cz, jan.zeman@cvut.cz}}

\begin{document}

\maketitle
\thispagestyle{empty}
\pagestyle{empty}

\begin{abstract}
The emerging field of passive macro-scale tile-based self-assembly (TBSA) shows promise in enabling effective manufacturing processes by harnessing TBSA’s intrinsic parallelism. However, current TBSA methodologies still do not fulfill their potentials, largely because such assemblies are often prone to errors, and the size of an individual assembly is limited due to insufficient mechanical stability. Moreover, the instability issue worsens as assemblies grow in size. Using a novel type of magnetically-bonded tiles carried by bristle-bot drives, we propose here a framework that reverses this tendency; i.e., as an assembly grows, it becomes more stable. Stability is achieved by introducing two sets of tiles that move in opposite directions, thus zeroing the assembly net force. Using physics-based computational experiments, we compare the performance of the proposed approach with the common orbital shaking method, proving that the proposed system of tiles indeed possesses self-stabilizing characteristics. Our approach enables assemblies containing hundreds of tiles to be built, while the shaking approach is inherently limited to a few tens of tiles. Our results indicate that one of the primary limitations of mechanical, agitation-based TBSA approaches, instability, might be overcome by employing a swarm of free-running, sensorless mobile robots, herein represented by passive tiles at the macroscopic scale.
\end{abstract}



\section{Introduction and state-of-the-art}

Tile-Based Self-Assembly (TBSA) is a process in which building units (tiles), starting from an initial disordered state, organize into a target pattern. The process itself is passive, being driven by predefined interaction rules among tiles and by an external excitation affecting all tiles in an assembly. These simple yet universal features are making self-assembly processes a topic of investigation across multiple length scales and disciplines, e.g.,~\cite[and references therein]{Whitesides2002}. We concentrate on the most relevant processes relevant to our contributions below. 

\subsection{DNA-based self-assembly}

The most substantial progress in TBSA processes to date has been achieved in DNA-based algorithmic self-assembly at the nanoscale, established in a pioneering work of Winfree~\cite{winfree_algorithmic_1998}. In the Winfree's framework, individual tiles are produced from DNA strands, with inter-tile interactions relying on the chemical bonds between their edges, and the overall process controlled by temperature. This direction of inquiry has witnessed remarkable growth, from 7 nm DNA cubes~\cite{Chen1991} to Sierpinski triangle~\cite{SierpinskiDNAAssembly} assemblies, bitmaps formed from 8,704 molecular pixels~\cite{Tikhomirov2017}, up to robust molecular computations with error rates as low as 1:3,000~\cite{Woods2019}. 

These achievements were all made possible by concurrent development of modeling strategies, with Winfree's abstract Tile Assembly Model (aTAM) providing the fundamental formalization (see Section~\ref{sec:aTAM_def} for a brief overview). Despite many adopted simplifications, aTAM has proved helpful in, e.g., rigorous analyses of TBSA~\cite{Soloveichik_complexity} or in the optimal design of tiles and their interactions~\cite{Ma_PATS,Goos_PSH} for assemblies growing from a~given initial (seed) structure.  

\subsection{Centimeter-scale, magnetically-based self-assembly}

Upscaling the principles and results of DNA-based self-assembly to the macroscale has been motivated by the development of non-conventional, intrinsically parallel, fabrication techniques for architectured materials and components~\cite{Boncheva2005}. Most attempts to date have relied upon magnetic forces to encode tile interactions, e.g.,~\cite{majumder_framework_2008,daiko_tsutsumi_multistate_2007,masumori_morphological_2013}, employing mechanical excitation~\cite{daiko_tsutsumi_multistate_2007,masumori_morphological_2013,CentimeterscaledSA, Ipparthi_2018}, fans~\cite{majumder_framework_2008}, fluid turbulence~\cite{Hageman_2018,Lothman_2019,Hafez2021}, or temperature~\cite{Han2022} to guide the self-assembly process. 

So far, the complexity of patterns achievable at the centimeter scale has fallen short of nanoscale DNA-based results. For instance, our initial macroscale attempt~\cite{CentimeterscaledSA} resulted in $2 \times 2$ square patterns or up to 7-tile long linear chains, whereas a recent work by Han \emph{et al.}~\cite{Han2022} yielded similarly-sized chains that disassembled as temperature changed. Tsutsumi and Murata~\cite{daiko_tsutsumi_multistate_2007} succeeded in assembling a 10-tile Sierpinski triangle, using specifically tailored inter-tile interaction, whereas Hafez~\emph{et al.}~\cite{Hafez2021} demonstrated the self-assembly of elementary two- and three-dimensional shapes. Our latest work~\cite{ral_21} reported on an errorless assembly of a $4 \times 4$ checkerboard pattern with carefully designed mechano-magnetic tile interactions. 

However, follow-up experiments have revealed that macroscale TBSA eventually reaches a critical size, where collisions with moving tiles and external excitation lead to detachment from the seed or assembly disintegration, cf. Fig. 12 (middle) in~\cite[p. 6]{ral_21}. This mechanical instability presents a fundamental obstacle to assembling the large and complex patterns necessary for real-world applications.

\section{Contributions}\label{sec:contributions}

This paper identifies the main parameters governing the phenomenon of assembly detachment and, leveraging this insight, proposes a framework that reverses the tendency towards instability at the macroscale with a self-assembled structure that is more stable as it grows in size. Particular emphasis is given to:

\begin{itemize} 
    \item Clarifying the intrinsic limit on the size of an assembly with a suitable scaling argument,
    \item Presenting a strategy for overcoming this limit by assuming that the motion of the tiles follows a generalized unicycle model with two distinct tile families,
    \item Proposing a framework for the physical realization of the system, utilizing a bristle-bot-like drive, e.g.,~\cite{Cicconofri2015233,bbot_steer,forward_backward}, that exhibits both translations and rotations,
    \item Demonstrating the self-stabilizing behavior of the novel system by extensive, real-time, physics-based simulations, with results being reproducible using the tools and code available in our public GitLab repository~(\url{https://gitlab.ciirc.cvut.cz/imr/expro/tiledyn2}).
\end{itemize}

To these goals, the paper is structured as follows. Section~\ref{sec:issue} briefly introduces the TBSA theory and provides insight into the instability phenomena occurring during TBSA. Section~\ref{sec:methodology} describes the solution we propose and introduces our macroscale TBSA simulator together with the simulations we performed to confirm the self-stabilizing effect of our novel TBSA framework. Section~\ref{sec:experiments} collects simulation-based results to compare chessboard patterns obtained by the commonly-used in-plane shaking approach to our proposed self-stabilizing approach. Section~\ref{sec:conclusion} discusses our findings; their future extensions and application scenario are addressed in Section~\ref{sec:future_work}. 

\section{Limitations of macroscale TBSA}\label{sec:issue}

\subsection{Abstract Tile Assembly Model}\label{sec:aTAM_def}

\begin{figure}[h]
    \centering
    \includegraphics[width=0.5\columnwidth]{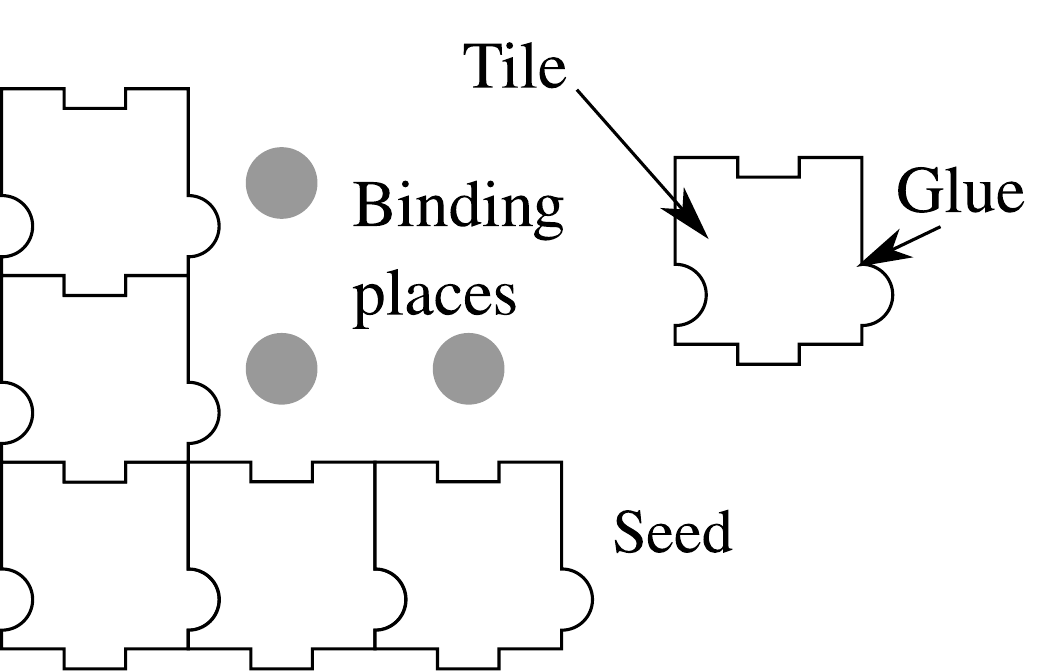}
    \caption{Illustration of the terminology used in this work. Adapted from~\cite{ral_21}.}
    \label{im:atam_example}
\end{figure}

Macroscopic tile-based self-assembly (TBSA) has not yet been formalized; therefore, the basic assumptions and principles of this study are inspired by Winfree's aTAM framework~\cite{winfree_algorithmic_1998}, in which self-assembling elements (\emph{tiles}) are squares, and each edge of a tile carries a \emph{glue}. Tiles and glues are passive and time-invariant. Glues can consist of different \emph{types}, with different \emph{strengths}. If \emph{matching} glues come together, the tiles carrying them are bonded (e.g., magnetically), and the assembly grows. Growth can be initiated only from a \emph{seed}. The tiles in an assembly can only occupy positions determined by a regular orthogonal grid---\emph{binding places}---and tiles are not allowed to rotate. This terminology is illustrated in Fig.~\ref{im:atam_example}.

The whole tile system is placed in a \emph{reactor} that supplies energy to the system. The magnitude of agitation is quantified by a positive integer value---\emph{temperature}---specifying the minimum number of matching glues needed to stably attach a free tile to an existing assembly.

\subsection{Assembly detachment}

\begin{figure}[h]
    \centering
    \includegraphics[width=0.72\columnwidth]{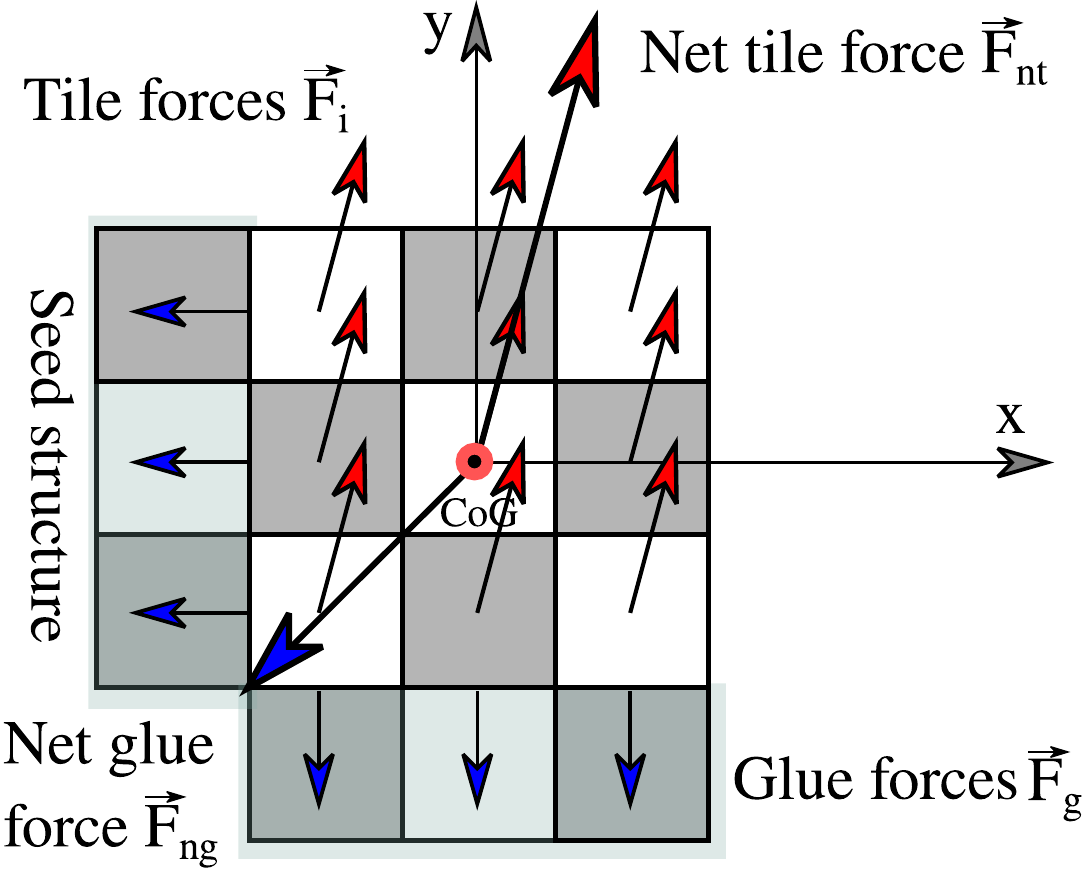}
    \caption{Configuration of tile forces leading to detachment or disintegration of the assembly.}
    \label{fig:stability_framework}
\end{figure}

To study the phenomenon of assembly detachment, we assume that the side of an L-shaped seed contains $n_\mathrm{s}$ tiles. The maximal assembly size is thus $n^2_\mathrm{s}$. The seed structure holds the entire assembly with the combined force of its $2n_\mathrm{s}$ glues, $\vec{F}_\mathrm{ng}$ (net glue force); see Fig.~\ref{fig:stability_framework}. All the tiles are under the influence of an additional driving force $\vec{F}_\mathrm{nt}$ (the net tile force). Assuming that the $i$-th tile is affected by the driving force $\vec{F}_i$, the net tile force exerted on the connection between the L-shaped seed and the assembly becomes
\begin{equation}
    \vec{F}_\mathrm{nt} 
    = 
    \sum_{i=1}^{n^2_\mathrm{s}} \vec{F}_i.
\end{equation}
Neglecting friction among tiles themselves and tiles and a reactor floor, the assembly detaches when
\begin{equation}
    \vec{F}_\mathrm{nt}+\vec{F}_\mathrm{ng} \text{ does not occupy the $3^{rd}$ quadrant,}
    \label{eq:detachment}
\end{equation}
assuming a right-handed coordinate system with an origin in the Center of Gravity~(CoG). 

The detachment condition is easily satisfied in shaking-based TBSA. Here, the reactor moves orbitally along a trajectory that allows free tiles to attach to a growing assembly and wrongly connected tiles to disconnect. Because all tiles are experiencing the same acceleration $\vec{a}_\mathrm{nt}$, the net force becomes 
\begin{equation}\label{eq:Fnt}
    \vec{F}_\mathrm{nt} 
    = n_\mathrm{s}^2 m_\mathrm{t} \vec{a}_\mathrm{nt},
\end{equation}
while the net glue force can be expressed as (recall Fig.~\ref{fig:stability_framework})
\begin{equation}\label{eq:Fng}
\vec{F}_\mathrm{ng} 
= 
2n_\mathrm{s} F_\mathrm{g} \frac{\sqrt[]{2}}{2} \left[1, 1 \right]^{\mathrm{T}}.
\end{equation}
In Eq.~\eqref{eq:Fnt}, $m_\mathrm{t}$ represents the mass of a single tile and $F_\mathrm{g}$ in Eq.~\eqref{eq:Fng} the magnitude of a single glue force.

Because the reactor motion is harmonic, the detachment condition in Eq.~\eqref{eq:detachment} becomes equivalent to 
\begin{equation}
\| \vec{F}_\mathrm{ng} \| < \| \vec{F}_\mathrm{nt} \|,
\end{equation}
with $\| \cdot \|$ denoting the Euclidean norm. Employing Eqs.~\eqref{eq:Fnt} and~\eqref{eq:Fng}, we recast the condition to 
\begin{align}\label{eq:fund_limit}
n_\mathrm{s} 
>
\sqrt{%
\frac{ 2F_\mathrm{g} }{m_\mathrm{t} \| \vec{a}_\mathrm{nt} \|}
},
\end{align}
thereby revealing the fundamental limit on the assembly size.

\section{Methodology}\label{sec:methodology}

This section presents our strategy for tackling the instability issue inherent in seeded self-assembly, organized into three parts for the readers' convenience. The first part (Section~\ref{ssec:principles}) introduces the general strategy for achieving the self-stabilizing property. The second part outlines the experiment setup (Section~\ref{sec:outline_realization}) and presents the details on a physics-based model for self-stabilizing self-assembly (Section~\ref{sec:phys_model}). The third part describes the simulation software (Section~\ref{sec:physics_solver}) as well as data analysis techniques (Section~\ref{sec:data_analysis}). 

\subsection{Self-stabilizing TBSA principle}\label{ssec:principles}

Section~\ref{sec:issue} revealed that reducing net tile force $\vec{F}_\mathrm{nt}$ is the key to increasing the assembly size. Eq.~\eqref{eq:fund_limit} suggests that this increase can be achieved by decreasing the magnitude of the acceleration of agitation, $\| \vec{a}_\mathrm{nt} \|$. Such an approach is of limited use because decreased agitation leads to deceleration of assembly and higher error rates~\cite{ral_21}. 

\begin{figure}[h]
    \centering
    \includegraphics[width=0.95\columnwidth]{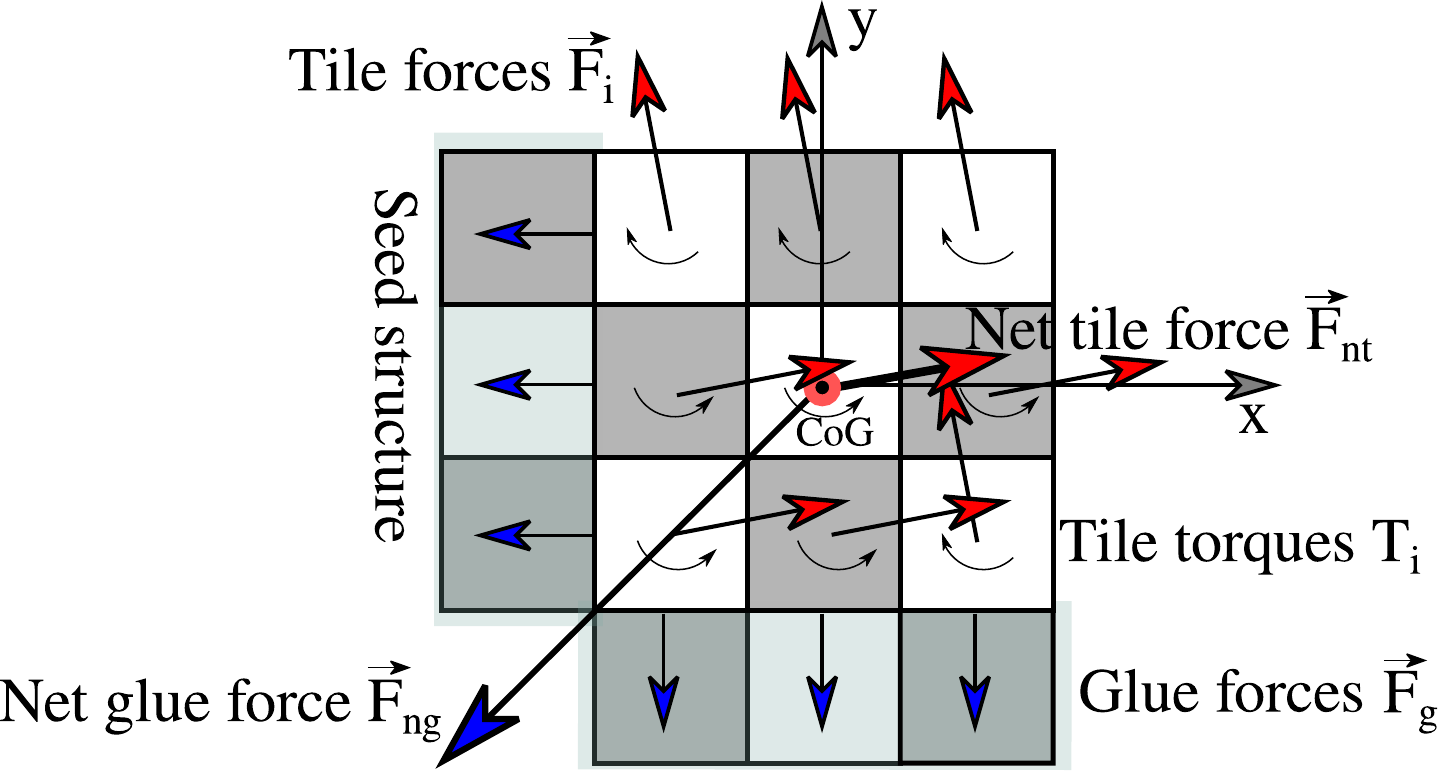}
    \caption{Configuration of tile forces resulting in minimization of the net force.}
    \label{fig:minimization}
\end{figure}

The proposed solution is more complex and results in a zeroing of the net force on the tiles within the assembly, with minimal impact on the assembly process.

To minimize the net tile force $\vec{F}_\mathrm{nt}$ directly, we propose a TBSA system that incorporates not only tile types, glues, seed, and temperature, but also a set of rules describing tile movements during the assembly through time. This system employs a circular reactor with a cross-shaped seed fixed in the center (an L-shaped seed in a square reactor would also be permissible), and we consider that the motion of the freely moving $i$-th tile follows an extended unicycle model~\cite{Aicardi} 
\begin{subequations}\label{eq:mmodel}
\begin{align}
    \dot{x}_i(t) & = u(t)a_i \cos(\phi_i t) + x_{\mathrm{d}i}(t), \\
    \dot{y}_i(t) & = u(t)b_i \sin(\phi_i t) + y_{\mathrm{d}i}(t), \\
    \dot{\phi}_i(t) & = u(t)\omega_i + \phi_{\mathrm{d}i}(t). 
\end{align}
\end{subequations}
Here $x_i$, $y_i$, and $\phi_i$ represent the rigid-body translations and rotations of the $i$-th tile, the superimposed dot represents the time derivative; $a_i$, $b_i$, and $\omega_i$ are $i$-th tile parameters; $x_{\mathrm{d}i}$, $y_{\mathrm{d}i}$, and $\phi_{\mathrm{d}i}$ represent inter-tile interactions; and $u(t) \in [ -1, 1 ]$ is a periodic function accounting for external excitation.

Moreover, we assume that the tiles are grouped into two distinct families, each sharing the same set of parameters $a_i, b_i$, and $\omega_i$. At the same time, the parameters for both tile families have opposite signs, so their motion occurs in opposite directions. 

Now, assuming that both tile families contain the same number of tiles, the net tile force $\vec{F}_\mathrm{nt}$ will converge to zero as the assembly grows. On the other hand, the net glue force $\vec{F}_\mathrm{ng}$ will increase in time as more tiles are connected to the seed, recall Eq.~\eqref{eq:Fng}. Thus, the resultant $\vec{F}_\mathrm{nt} + \vec{F}_\mathrm{ng}$ will satisfy the condition given by Eq.~\eqref{eq:detachment} as the size of the assembly increases. Note that the torques depicted in Fig.~\ref{fig:minimization} do not introduce any (de)stabilization effects because their sum converges to zero. Still, they are necessary so that the free tiles follow the unicycle kinematics that results in more mixing of free tiles.

\begin{figure}
	\centering
	\vspace*{0.2cm}
	\includegraphics[width=0.85\columnwidth]{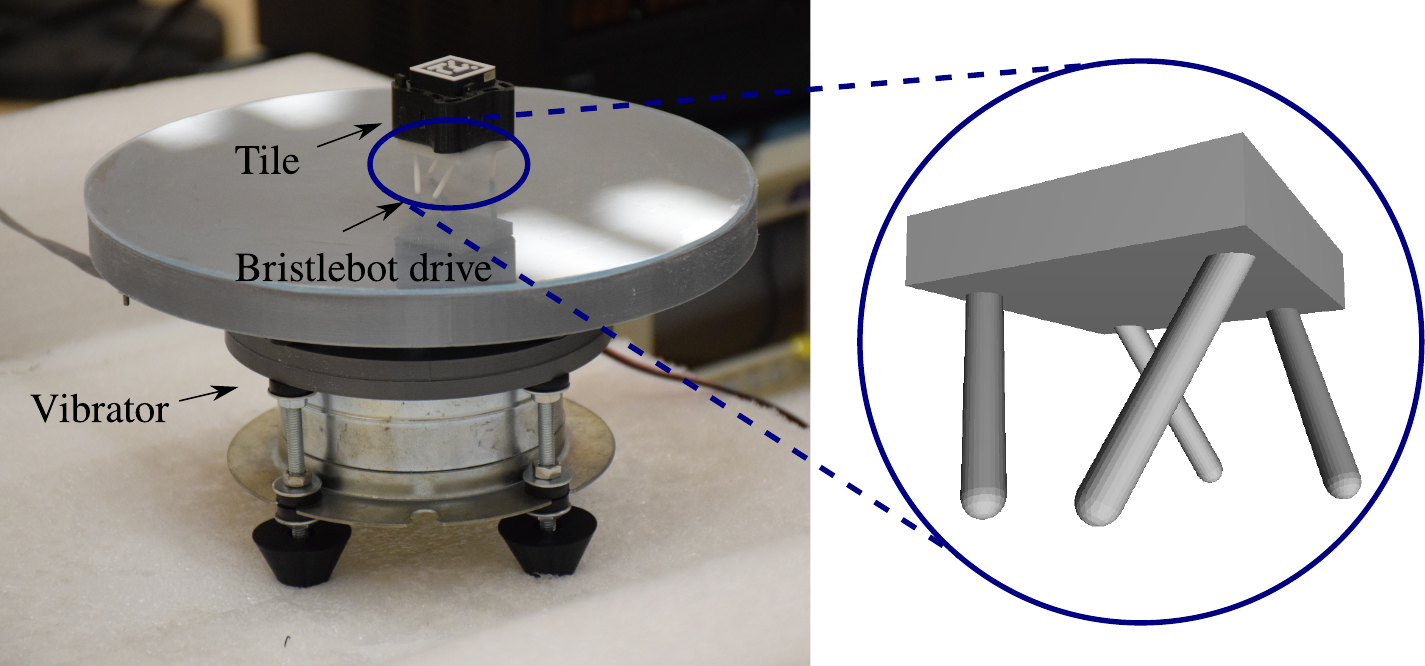}
	\caption{Illustration of a tile with a bristle-bot drive placed onto a vertical shaker.}
	\label{fig:bbot_drive}
\end{figure}

\begin{figure}
	\centering
	\vspace*{0.3cm}
	\includegraphics[width=0.95\columnwidth]{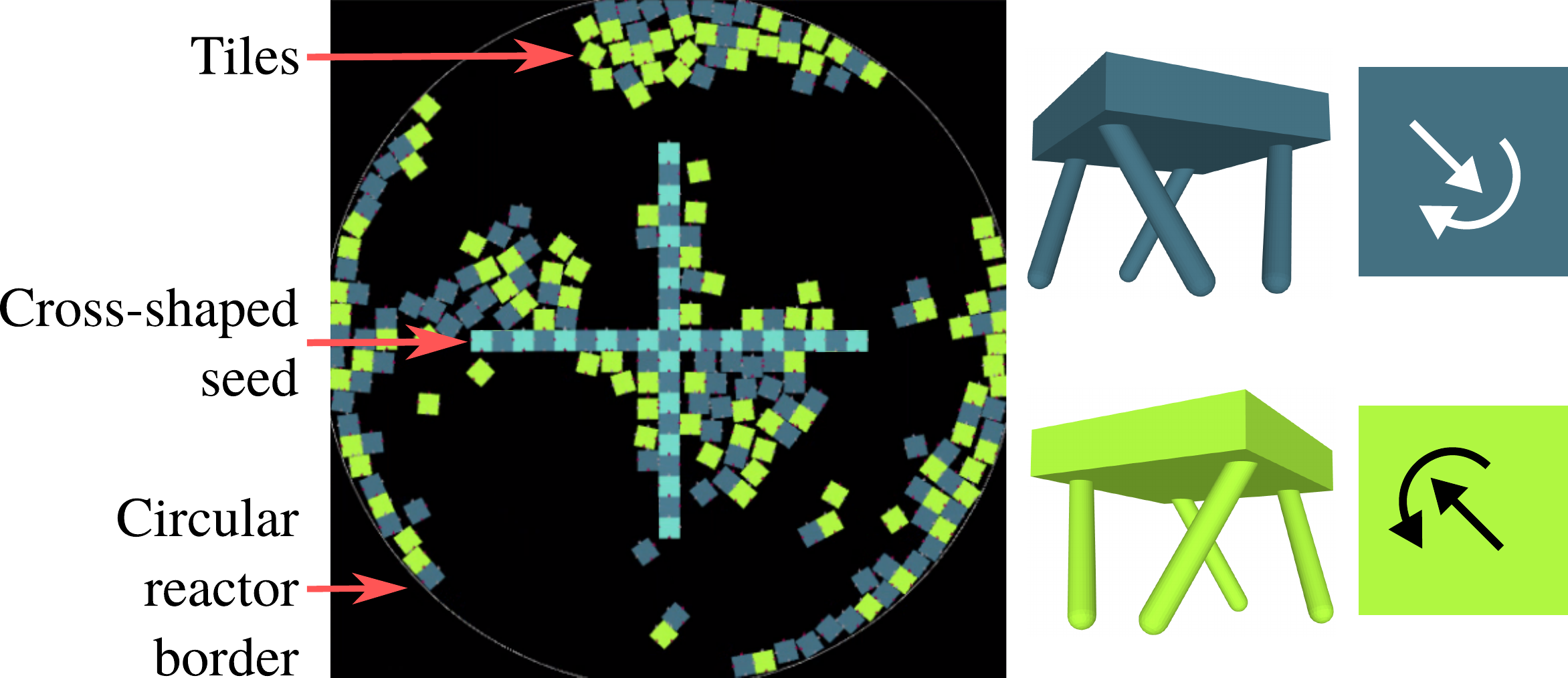}
	\caption{Physical setup of the proposed self-stabilizing framework with the corresponding bristle-bot drives and the forward directions of motion indicated for both tile families (direction of motion can be switched to backward by changing the actuation mode).}
	\label{fig:self_stabilizing}
\end{figure}

\subsection{Outline of practical realization}\label{sec:outline_realization}

Such self-stabilizing behavior can be reached in practice by introducing two types of tile \emph{drives} (representing two tile families), each exerting a force in the opposite direction. This can be achieved by combining each tile with a bristle-bot drive, shown in Fig.~\ref{fig:bbot_drive}. Bristle-bots are simple rigid-bodied robots with bristles on their lower parts, see Fig.~\ref{fig:bbot_drive}. The bristles are placed such that they give each robot a preferred direction of motion (linear or even curved~\cite{bbot_steer}) when actuated by vertical vibrations either in the robot body or on the surface under the robot~\cite{Cicconofri2015233}. Moreover, the direction of motion can be controlled and even reversed by changing the actuation mode (vibration amplitude and frequency)~\cite{forward_backward}.

Therefore, our proposed setup involves a flat vibrating surface excited by a shaker actuated by a harmonic signal with controllable amplitude and frequency, illustrated in Fig.~\ref{fig:bbot_drive}. The individual bristles under the tile are oriented asymmetrically, see Fig.~\ref{fig:bbot_drive}, to allow the bristle-bots to exhibit both translation and steering movements switchable according to the excitation signal, in agreement with the unicycle model Eq.~\eqref{eq:mmodel}. The self-assembly process takes place in a circular reactor and is initiated starting from a cross-shaped seed, illustrated in Fig.~\ref{fig:self_stabilizing}.

\subsection{Physics-based self-assembly model}\label{sec:phys_model}

Following from the preceding discussion, the minimal physics-based model needs to incorporate three components: a tile body, inter-tile interactions representing glues, and a tile drive describing the external excitation. These are discussed directly below in more detail. 

Note that this paper aims neither to exactly reproduce the motion of bristle-bots nor to explore the behavior of the bristle-bots. Instead, we focus on comparing the self-assembly mechanisms from the proposed framework and our earlier results~\cite{ral_21}, using the idealized model.

\subsubsection{Tile body}

\begin{figure}[h]
    \centering
    \includegraphics[width=0.6\columnwidth]{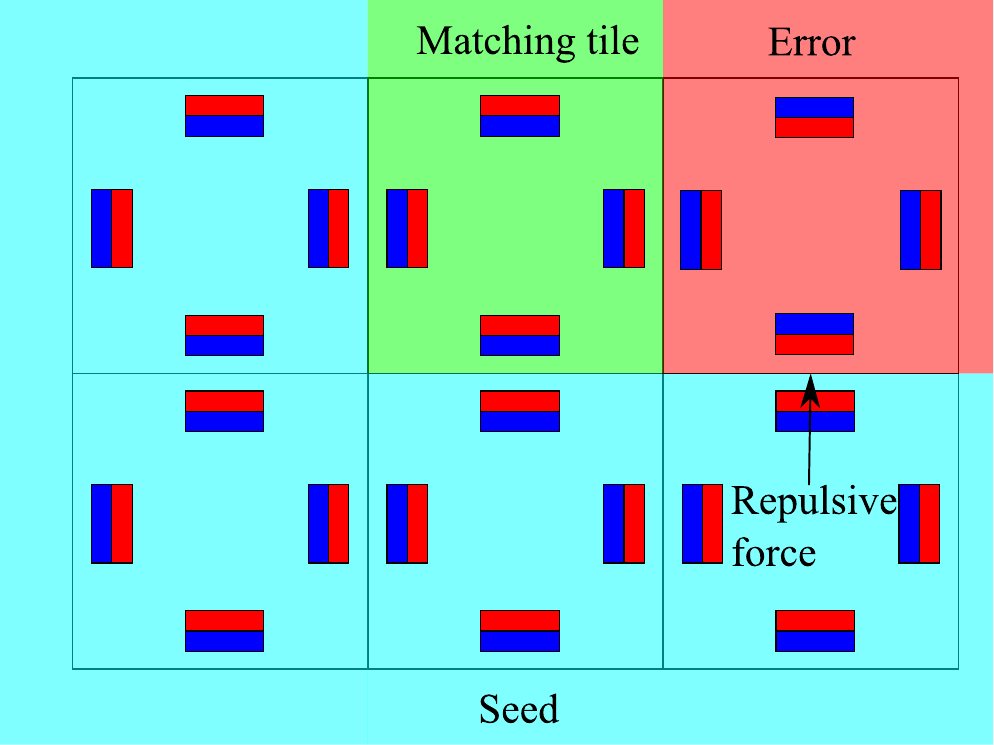}
    \caption{Illustration of placement of magnets within tiles and their function for building an assembly. View from the top.}
    \label{fig:tile_magnets}
\end{figure}

Tiles are represented by rigid squares, and the Coulomb friction model is adopted to account for frictional effects between tile edges and between the tiles and the surface upon which they move. Each tile carries glue in the center of each edge representing a small neodymium magnet. Magnets are placed so that their repulsive forces prevent a tile from bonding to the seed in the wrong orientation. An assembly illustrating the placement of magnets is in Fig.~\ref{fig:tile_magnets}.

The mechanical model of a tile in the simulation solver is characterized by the parameters listed in Table~\ref{tab:mech_model} (the parameters were obtained by fitting simple experiments). We used the glues based on the abstract magnetic model~\cite{CentimeterscaledSA}, and, for programming purposes, we labeled the attracting pairs of glues with numbers having opposite signs, i.e., $\{+1, -1\}, \{+2, -2\}$, and so on.

\begin{table}[h]
	\vspace*{0.15cm}
    \caption{Tile parameters set in the simulation solver.}
    \begin{center}
        \begin{tabular}{r|c}
            Parameter & Value \\
            Tile restitution & 0.2 \\
            Linear friction coefficient (tile-tile) & 0.25 \\
            Linear friction coefficient (tile-floor) & 0.25 \\
            Angular friction coefficient (tile-floor) & 0.25 \\
            Tile mass & 16 g \\
            Tile width & 3 cm \\
            Linear damping coefficient & 0.8 \\
            Angular damping coefficient & 0.5
        \end{tabular}
    \end{center}
\label{tab:mech_model}
\end{table}

\subsubsection{Magnetic force model}

\begin{figure}
    \centering
    \includegraphics[width=0.9\columnwidth]{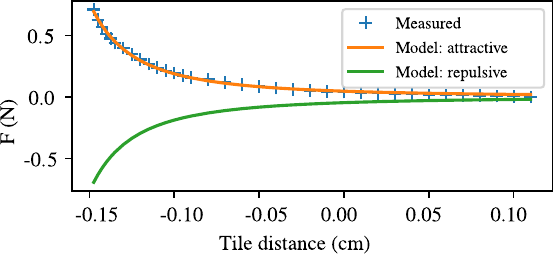}
    \caption{Force between two aligned magnets with respect to their distance. The magnets in each tile are offset 0.15 cm into the tile body. This corresponds to reality because when two tiles bind, their edges (together with magnets) are aligned in the same orientation.}
    \label{fig:magnetic_force}
\end{figure}

The magnetic field between tiles can be determined using Maxwell equations. However, their solution is computationally expensive when our model runs the near- to real-time performance required for long experiments. The dimensions of the magnets are relatively small compared to tiles, and their force is close to zero for a distance equal to one tile dimension. Thus, three or more magnets cannot get into close contact and interact in any meaningful way. Therefore, our model considers only sufficiently close magnets, and the interaction force between all pairs of glues $(g_i, g_j)$ in the system is approximated as:
\begin{equation}
\vec{F}(\vec{r}_i, \vec{r}_j) = p\frac{\alpha}{(\|\vec{r}_j-\vec{r}_i\|+\beta)^2}\frac{\vec{r}_j-\vec{r}_i}{\|\vec{r}_j-\vec{r}_i\|}
\label{eq:mag_model}
\end{equation}
where $r_i$ and $r_j$ define the positions of magnet centers, the parameter $p$ equals $1$ when the respective glue magnets are in an orientation resulting in attractive force; otherwise $p$ equals $-1$. Parameters $\alpha = 0.18$ Nm$^2$ and $\beta = 0.64$ m were determined from fitting the data obtained by measurement of real magnet forces, see Fig.~\ref{fig:magnetic_force}, using the non-linear least-squares method.

\subsubsection{Tile drive}
Only the tiles forming the cross-shaped seed, see Fig. \ref{fig:self_stabilizing}, are static (lacking the drive), while the others are driven by force $\vec{F}_i$. We do not model the full bristle-bot drive dynamics, but rather replace this with a combination of force acting in the center of a tile, and a torque, allowing the tile to rotate, thereby allowing universal control of the motion of a tile. To compare the proposed self-stabilizing strategy to macroscale self-assembly and the common strategy presented in our previous work~\cite[and the references therein]{ral_21}, we implemented two types of motion simulating the unicycle model and a reactor shaking in a horizontal plane.

The unicycle-like dynamics of the $i$-th tile is achieved by application of force $\vec{F}_i$ given as:
\begin{equation}
    \vec{F}_i(t) = u(t)a_iF_\mathrm{mag}\left[\sin(\phi_i), \cos(\phi_i)\right]^{\rm{T}},
\end{equation}
where $u(t) = \sin(2\pi f t)$~(with $f$ representing the motion frequency), $F_\mathrm{mag}$ is a constant controlling the magnitude of the force, and $\phi_i$ is the orientation of a tile, recall Eq.~\eqref{eq:mmodel}. In this case, frequency $f$ was set to $0.1$~Hz meaning that the direction of motion was reversed every $10$~s. 

The torque around the $z$-axis is given by
\begin{equation}
    T(t) = u(t)\omega_iT_\mathrm{mag},
\end{equation}
where $T_\mathrm{mag}$ is a constant controlling the magnitude of torque.

The shaking-based motion model does not apply any additional torque on tiles, and the force applied on a tile is taken as 
\begin{equation}
    \vec{F}_i(t) = F_\mathrm{mag}\left[ \sin(2\pi f t), \cos(2\pi f t) \right]^{\rm{T}}
    \label{eq:shaking_model}
\end{equation}
with $f=0.33$~Hz (the reactor circumscribes its circular trajectory every $3$~s).

\subsection{Simulation solver and its implementation}\label{sec:physics_solver}
\begin{figure}[h]
    \centering
    \vspace*{0.3cm}
    \includegraphics[width=0.9\columnwidth]{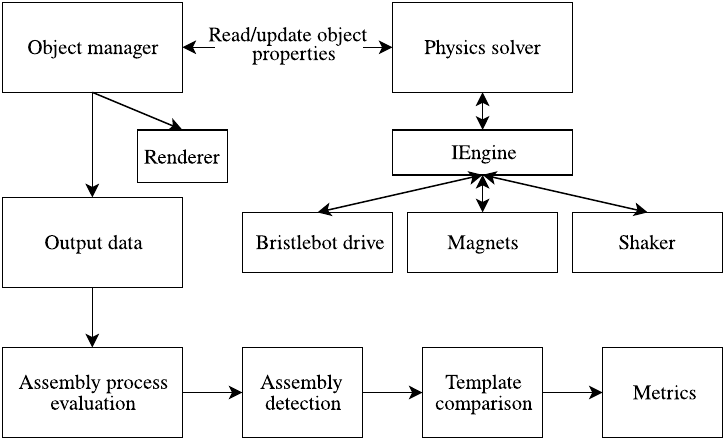}
    \caption{Simulation software architecture.}
    \label{fig:sim_arch}
\end{figure}

We used the Box2D real-time physics engine (\url{https://box2d.org}) to solve the equations of motion of the system.\footnote{%
The 2D engine was chosen after a previous experience with a 3D solver Bullet~(\url{https://pybullet.org}) which was unable to guarantee an~artifact-free motion of tiles in two dimensions. Moreover, the 2D engine offers the required performance and numerical stability. 
}
The implementation resulted in a single-purpose software that can simulate TBSA experiments on a physical level. Its architecture is described in Fig.~\ref{fig:sim_arch}. The software can be extended via engines---modules that take the simulation state and output forces and torques into account. We used this functionality to implement the unicycle and shaking-based motion models and the magnetic interaction between tiles. The software features not only a physics engine but also an OpenGL renderer. The data collected can be analyzed offline.

\subsection{Data analysis}\label{sec:data_analysis}
The simulator was programmed to save data on an assembly state for every 10 s of the running simulation. Data were subsequently post-processed following four steps:
\begin{enumerate}
    \item Detection of tiles which belong to the seed assembly,
    \item Detection of assembly holes,
    \item Comparison of the actual pattern with a ground truth,
    \item Detection of errors.
\end{enumerate}

The three metrics we collected and analyzed were: the size of an assembly (including errors), percentage of errors (i.e., number of erroneously placed tiles related to the number of tiles in the assembly at a given time), and the percentage of holes (i.e., number of empty spaces in the assembly surrounded by four tiles related to the number of tiles in the assembly at a given time).

\section{Results} \label{sec:experiments}

The main goal of the experiments performed was a comparison of assembly performance using both the shaking-based and our proposed approaches. We focused especially on achievable assembly size, number of errors and the tendency to form holes. 

\begin{figure}[h]
	\begin{center}
		\vspace*{0.2cm}
		\includegraphics[width=0.75\columnwidth]{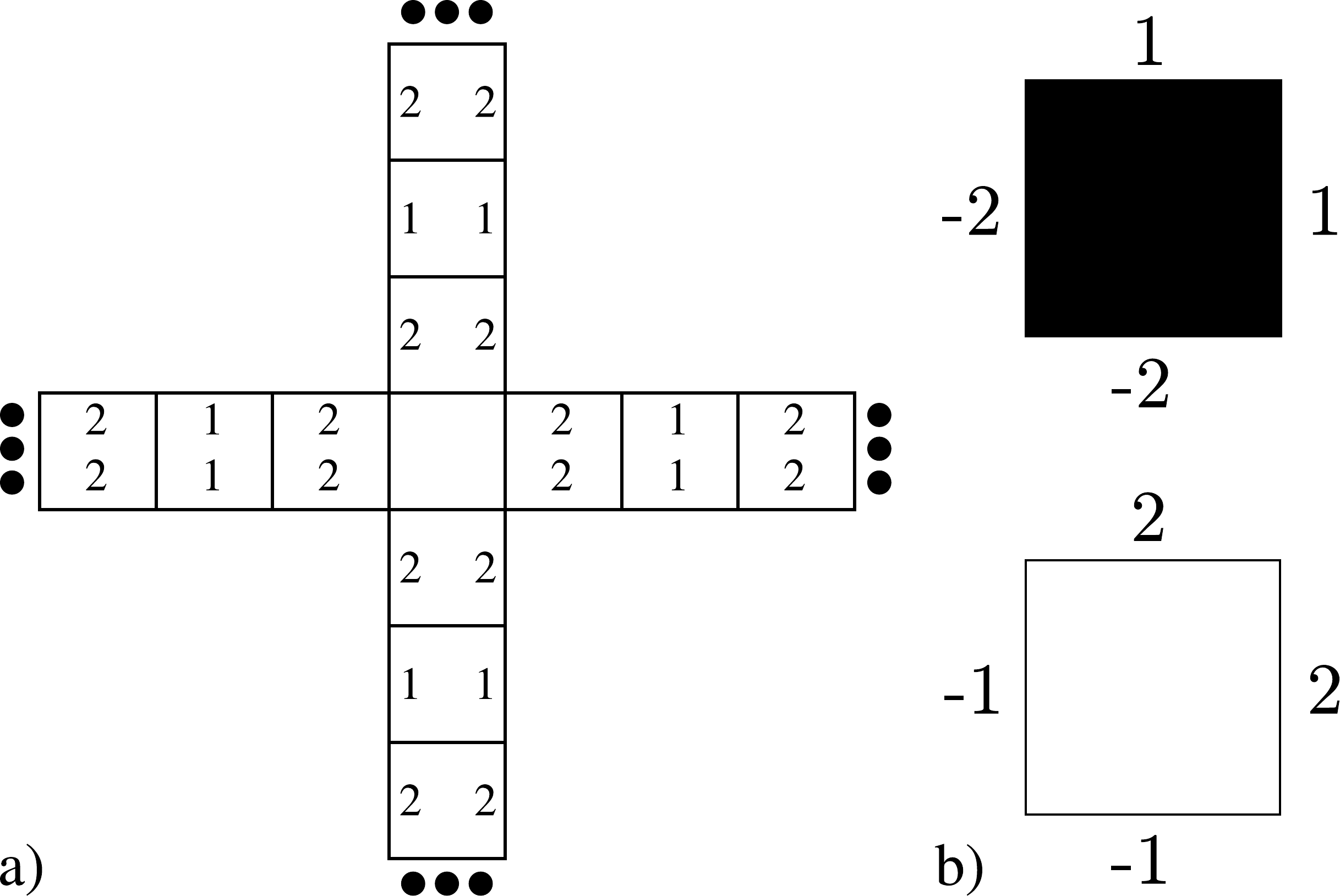}
		\caption{High-level overview of a chessboard assembling tileset, a) cross-shaped seed for a chessboard assembly, b) chessboard assembling tileset.}
		\label{fig:chess_tileset}
	\end{center}
\end{figure}

We performed a chessboard assembly, which required two types of tiles (black and white) and two types of tile bonds leading to four types of glues: $\pm1$ and $\pm2$, see Fig.~\ref{fig:chess_tileset}. We placed a cross-shaped seed with a size of $10\times 10$ tiles and $550$ free tiles into a reactor with a radius of $60$ cm.
First, we ran experiments with the shaking-based reactor model, i.e., force acted on the free tiles with orbital movements of the reactor in a horizontal plane. This model approximated the real, orbitally-shaken reactor described and used in our previous work \cite{ral_21}, where the reactor body moved along a circular trajectory. Thus, $\vec{F}_i$ for all the tiles $T_i$ had the same orientation and magnitude, following Eq.~\eqref{eq:shaking_model}. Second, we ran experiments with a periodic harmonic signal to drive the unicycle motion model as a simulation of the proposed self-stabilizing approach, where the forces acting on the tiles had different sizes and directions.

\begin{figure}[h]
\begin{center}
    \includegraphics[width=0.95\columnwidth]{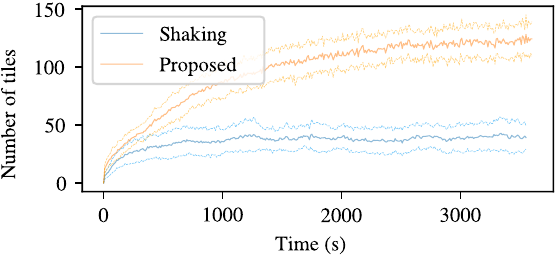}
    \caption{Snapshots from a 3,600 s simulation of a chessboard assembly experiment using the proposed self-stabilizing approach (unicycle motion model).}
    \label{fig:self-stab_simulation}
\end{center}
\end{figure}

\begin{figure}[h]
    \begin{center}
    	\vspace*{0.2cm}
        \includegraphics[width=0.95\columnwidth]{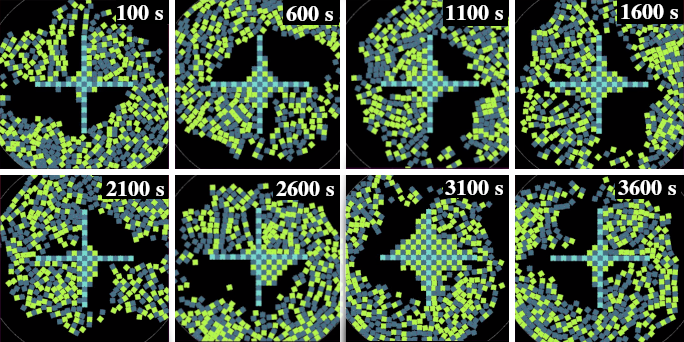}
        \caption{Snapshots from a 3,600 s simulation of a chessboard assembly experiment using the shaking-based approach.}
    \label{fig:shaking_simulation}
    \end{center}
\end{figure}

\begin{figure}[h]
	\includegraphics[width=0.98\columnwidth]{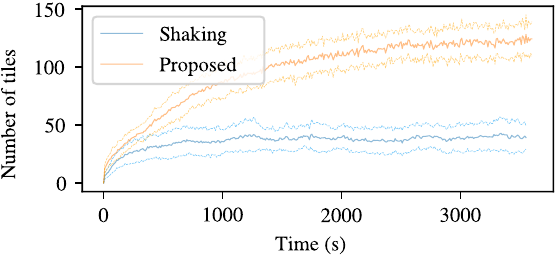}
	\caption{Evolution of the average assembly size and its standard deviation.}
	\label{fig:size_chess}
\end{figure}

After running $36$ simulated experiments for both shaking-based and self-stabilizing approaches \footnote{See the Multimedia material supplement for a video with a representative run for both motion models. Better quality video is also available at: \\ \url{https://youtu.be/75VPmPSLj3g}.}, we collected and analyzed the data with respect to the above-stated metrics. The assembly size reached by shaking in a horizontal plane was approximately $35$ tiles, while the self-stabilizing approach assembled more than $130$ tiles, illustrated in Figs.~\ref{fig:self-stab_simulation} and~\ref{fig:shaking_simulation}. Moreover, trends of the curves in Fig.~\ref{fig:size_chess} indicate that the proposed approach, in contrast to the shaking-based one, did not reach its limit, and the assembly would continue to grow if a higher number of tiles were available, albeit at a decreasing rate.

\begin{figure}[h]
	\vspace*{0.2cm}
	\includegraphics[width=0.96\columnwidth]{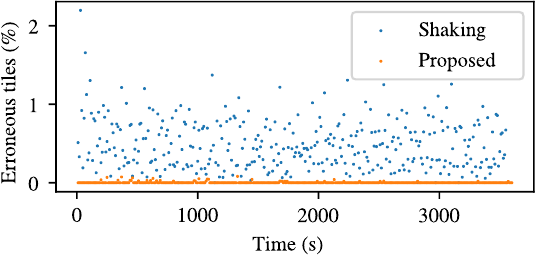}
	\caption{Evolution of the average percentage of errors in the assembly related to the number of tiles in the assembly.}
	\label{fig:nerrors_chess}
\end{figure}

\begin{figure}[h]
	\includegraphics[width=0.98\columnwidth]{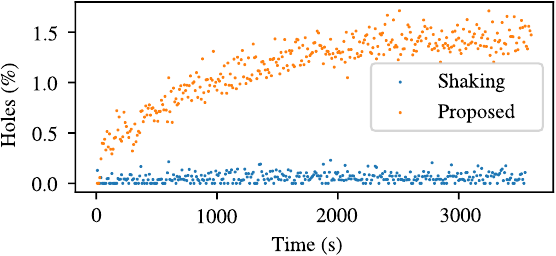}
	\caption{Evolution of the average percentage of holes in the assembly related to the number of tiles in the assembly.}
	\label{fig:nholes_chess}
\end{figure}

The average percentage of errors at the end of the experiment was significantly lower for the self-stabilizing approach, reaching $0 \%$, whereas the shaking-based approach reached $0.5 \%$, see Fig.~\ref{fig:nerrors_chess}, consistently with the values reported earlier in~\cite{ral_21}. On the other hand, the shaking-based approach provided a lower average percentage of holes at the end of experiment ($0 \%$) than the self-stabilizing one ($1.5 \%$), shown in Fig.~\ref{fig:nholes_chess}.

\section{Conclusions} \label{sec:conclusion}

We proposed here a novel self-stabilizing TBSA framework tackling the stability issue inherent to macroscale seeded self-assembly. The proposed approach assumes that tiles are divided into two distinct families, and the tiles in both families are subjected to the same magnitudes of force and torque but act in opposite directions, thus zeroing the assembly net force. Moreover, the direction of tiles is switchable from forward to backward, and switching is entirely controlled by a single parameter (amplitude or frequency). This introduces a tile motion model as a necessary part of the tilesystem description. 

To verify the functionality of the self-stabilizing framework, we implemented a physics-based simulation of the system, loosely based on differential bristle-bot kinematics, and compared the proposed approach to the state-of-the-art self-assembly shaking-based agitation approach. The experiments simulating the assembly of a chessboard pattern confirmed a significant difference in assembly performance in regard to error rates and maximum achievable sizes, with the self-stabilizing approach performing better than the shaking-based approach. This is because each assembly originating from the shaking-based self-assembly framework eventually reaches a limit when it detaches from the seed. An assembly created within the proposed self-stabilizing framework is not only unlimited in size, but its stability increases as the assembly grows and it had a lower percentage of erroneously placed tiles in our simulations. However, our proposed method resulted in a higher percentage of holes in the final assembly than the shaking-based framework. 

\section{Future work and potential applications}\label{sec:future_work}

The next step in our research will lead to physical realization of the proposed framework using a passive tile drive. The drive will consist of an undercarriage propelling magnetic~\cite{CentimeterscaledSA} or magneto-mechanical~\cite{ral_21} tiles we have already developed and tested. Since our preliminary experiments utilizing mechanically actuated bristle-bots indicated a lack of reliability and repeatability, our further research will move in the direction of actuation via a magnetic field~\cite{magnetic_bbot}.

Tackling the instability issue would enable, e.g., making a step towards automated manufacturing of modular architectured materials~\cite{Nezerka_2018,Tyburec_2022}, with the geometry of the modules and their placement optimized for target performance. In particular, we envisage the tiles as drives for molds representing individual modules. The tiles would self-assemble into a desired pattern, providing a mold for fabricating a complete product. Interested readers are referred to~\cite{Boncheva2005} for additional potential application scenarios.

\section{Acknowledgments}

The research leading to these results was funded by the Czech Science Foundation (GAČR) under grant agreement No. 19-26143X. Miroslav Kulich's work has been supported by the European Regional Development Fund under the Robotics for Industry 4.0 project (reg. no. CZ.02.1.01/0.0/0.0/15003/0000470). All authors thank anonymous referees and the editor for their constructive comments on the original version of the manuscript. The help of Dr. Stephanie Krueger, who provided editorial assistance, is also gratefully acknowledged.


\begin{thebibliography}{10}

\bibitem{Whitesides2002}
G.~M. Whitesides and B.~Grzybowski, ``{Self-assembly at all scales},''
  \emph{Science}, vol. 295, no. 5564, pp. 2418--2421, 2002.

\bibitem{winfree_algorithmic_1998}
E.~Winfree, ``Algorithmic self-assembly of {DNA},'' {PhD}. thesis, California
  Institute of Technology, 1998.

\bibitem{Chen1991}
J.~Chen and N.~C. Seeman, ``Synthesis from {DNA} of a molecule with the
  connectivity of a cube,'' \emph{Nature}, vol. 350, no. 6319, pp. 631--633,
  1991.

\bibitem{SierpinskiDNAAssembly}
P.~W.~K. Rothemund, N.~Papadakis, and E.~Winfree, ``Algorithmic self-assembly
  of {DNA} sierpinski triangles,'' \emph{{PLoS} Biology}, vol.~2, no.~12, p.
  e424, 2004.

\bibitem{Tikhomirov2017}
G.~Tikhomirov, P.~Petersen, and L.~Qian, ``Fractal assembly of micrometre-scale
  {DNA} origami arrays with arbitrary patterns,'' \emph{Nature}, vol. 552, no.
  7683, pp. 67--71, 2017.

\bibitem{Woods2019}
D.~Woods, D.~Doty, C.~Myhrvold, J.~Hui, F.~Zhou, P.~Yin, and E.~Winfree,
  ``Diverse and robust molecular algorithms using reprogrammable {DNA}
  self-assembly,'' \emph{Nature}, vol. 567, no. 7748, pp. 366--372, 2019.

\bibitem{Soloveichik_complexity}
D.~Soloveichik and E.~Winfree, ``Complexity of self-assembled shapes,'' in
  \emph{DNA Computing}, C.~Ferretti, G.~Mauri, and C.~Zandron, Eds.\hskip 1em
  plus 0.5em minus 0.4em\relax Berlin, Heidelberg: Springer Berlin Heidelberg,
  2005, pp. 344--354.

\bibitem{Ma_PATS}
X.~{Ma} and F.~{Lombardi}, ``Synthesis of tile sets for {DNA} self-assembly,''
  \emph{IEEE Transactions on Computer-Aided Design of Integrated Circuits and
  Systems}, vol.~27, no.~5, pp. 963--967, 2008.

\bibitem{Goos_PSH}
M.~Göös, T.~Lempiäinen, E.~Czeizler, and P.~Orponen, ``Search methods for
  tile sets in patterned {DNA} self-assembly,'' \emph{Journal of Computer and
  System Sciences}, vol.~80, no.~1, pp. 297 -- 319, 2014.

\bibitem{Boncheva2005}
M.~Boncheva and G.~M. Whitesides, ``Making things by self-assembly,'' \emph{MRS
  Bulletin}, vol.~30, no.~10, pp. 736--742, 2005.

\bibitem{majumder_framework_2008}
U.~Majumder and J.~H. Reif, ``A {framework} for {designing} {novel} {magnetic}
  {tiles} {capable} of {complex} {self}-assemblies,'' in \emph{Unconventional
  {Computing}}.\hskip 1em plus 0.5em minus 0.4em\relax Berlin, Heidelberg:
  Springer Berlin Heidelberg, 2008, pp. 129--145.

\bibitem{daiko_tsutsumi_multistate_2007}
D.~Tsutsumi and S.~Murata, ``Multistate part for mesoscale self-assembly,'' in
  \emph{{SICE} Annual Conference 2007}.\hskip 1em plus 0.5em minus 0.4em\relax
  {IEEE}, 2007.

\bibitem{masumori_morphological_2013}
A.~Masumori and H.~Tanaka, ``Morphological computation on two dimensional
  self-assembly system,'' in \emph{ACM SIGGRAPH 2013 Posters}, ser. SIGGRAPH
  '13.\hskip 1em plus 0.5em minus 0.4em\relax New York, New York, USA: ACM
  Press, 2013.

\bibitem{CentimeterscaledSA}
M.~J{\'i}lek, M.~Kulich, and L.~P\v{r}eu\v{c}il, ``Centimeter-scaled
  self-assembly: A preliminary study,'' in \emph{Proceedings of the 17th
  International Conference on Informatics in Control, Automation and Robotics -
  ICINCO}.\hskip 1em plus 0.5em minus 0.4em\relax SciTePress, 2020, pp.
  438--445.

\bibitem{Ipparthi_2018}
D.~Ipparthi, T.~A.~G. Hageman, N.~Cambier, M.~Sitti, M.~Dorigo, L.~Abelmann,
  and M.~Mastrangeli, ``Kinetics of orbitally shaken particles constrained to
  two dimensions,'' \emph{Physical Review E}, vol.~98, no.~4, 2018.

\bibitem{Hageman_2018}
T.~A.~G. Hageman, P.~A. Löthman, M.~Dirnberger, M.~C. Elwenspoek, A.~Manz, and
  L.~Abelmann, ``Macroscopic equivalence for microscopic motion in a turbulence
  driven three-dimensional self-assembly reactor,'' \emph{Journal of Applied
  Physics}, vol. 123, no.~2, p. 024901, 2018.

\bibitem{Lothman_2019}
P.~A. Löthman, T.~A.~G. Hageman, M.~C. Elwenspoek, G.~J.~M. Krijnen,
  M.~Mastrangeli, A.~Manz, and L.~Abelmann, ``A thermodynamic description of
  turbulence as a source of stochastic kinetic energy for {3D} self-assembly,''
  \emph{Advanced Materials Interfaces}, vol.~7, no.~5, p. 1900963, 2019.

\bibitem{Hafez2021}
A.~Hafez, Q.~Liu, and J.~C. Santamarina, ``{Self-assembly of millimeter-scale
  magnetic particles in suspension},'' \emph{Soft Matter}, vol.~17, no.~29, pp.
  6935--6941, 2021.

\bibitem{Han2022}
J.~Han, Q.~Lahondes, and S.~Miyashita, ``Size changing soft modules for
  temperature regulated self-assembly and self-disassembly,'' in \emph{2022
  IEEE 5th International Conference on Soft Robotics (RoboSoft)}, 2022, pp.
  461--466.

\bibitem{ral_21}
M.~Jílek, M.~Somr, M.~Kulich, J.~Zeman, and L.~Přeučil, ``Towards a passive
  self-assembling macroscale multi-robot system,'' \emph{IEEE Robotics and
  Automation Letters}, vol.~6, no.~4, pp. 7293--7300, 2021.

\bibitem{Cicconofri2015233}
G.~Cicconofri and A.~DeSimone, ``{Motility of a model bristle-bot: A
  theoretical analysis},'' \emph{International Journal of Non-Linear
  Mechanics}, vol.~76, pp. 233--239, 2015.

\bibitem{bbot_steer}
Z.~Hao, D.~Kim, A.~R. Mohazab, and A.~Ansari, ``Maneuver at micro scale:
  Steering by actuation frequency control in micro bristle robots,'' in
  \emph{2020 IEEE International Conference on Robotics and Automation (ICRA)},
  2020, pp. 10\,299--10\,304.

\bibitem{forward_backward}
D.~Kim, Z.~Hao, A.~Mohazab, and A.~Ansari, ``On the forward and backward motion
  of milli-bristlebots,'' \emph{International Journal of Non-Linear Mechanics},
  vol. 127, p. 103551, 2020.

\bibitem{Aicardi}
M.~Aicardi, G.~Cannata, G.~Casalino, and G.~Indiveri, ``On the stabilization of
  the unicycle model projecting a holonomic solution,'' in \emph{in 8th Int.
  Symposium on Robotics with Applications, ISORA 2000, Maui}, 2000, pp. 11--16.

\bibitem{magnetic_bbot}
D.~Kim, Z.~Hao, T.~H. Wang, and A.~Ansari, ``Magnetically-actuated micro-scale
  bristle-bots,'' in \emph{2020 International Conference on Manipulation,
  Automation and Robotics at Small Scales}, 2020, pp. 1--6.

\bibitem{Nezerka_2018}
V.~Ne{\v{z}}erka, M.~Somr, T.~Janda, J.~Vorel, M.~Do{\v{s}}k{\'{a}}{\v{r}},
  J.~Anto{\v{s}}, J.~Zeman, and J.~Nov{\'{a}}k, ``A jigsaw puzzle metamaterial
  concept,'' \emph{Composite Structures}, vol. 202, pp. 1275--1279, 2018.

\bibitem{Tyburec_2022}
M.~Tyburec, M.~Do{\v{s}}k{\'{a}}{\v{r}}, J.~Zeman, and M.~Kru{\v{z}}{\'{\i}}k,
  ``Modular-topology optimization of structures and mechanisms with free
  material design and clustering,'' \emph{Computer Methods in Applied Mechanics
  and Engineering}, vol. 395, p. 114977, 2022.

\end{thebibliography}
\end{document}